\pdfoutput=1

\documentclass[11pt]{article}

\usepackage[final]{acl}

\usepackage{times}
\usepackage{latexsym}
\usepackage{xspace}

\newcommand{\systemname}{{\sc PRAISE}\xspace}

\usepackage[T1]{fontenc}

\usepackage[utf8]{inputenc}

\usepackage{microtype}

\usepackage{inconsolata}
\usepackage{tikz}
\usetikzlibrary{shapes.geometric, arrows.meta, positioning, fit}

\usepackage{graphicx}

\usepackage{hyperref}

\title{\systemname: Enhancing Product Descriptions with LLM-Driven Structured Insights}

\author{
 \textbf{Adnan Qidwai\textsuperscript{1}\thanks{equal contribution \dag corresponding author}},
 \textbf{Srija Mukhopadhyay\textsuperscript{1}\footnotemark[1]},
 \textbf{Prerana Khatiwada\textsuperscript{2}\footnotemark[1]},
 \textbf{Dan Roth\textsuperscript{3}},
 \textbf{Vivek Gupta\textsuperscript{4\dag}} \\
\textsuperscript{1}IIIT Hyderabad, \textsuperscript{2}University of Delaware,
\textsuperscript{3}University of Pennsylvania
\textsuperscript{4}Arizona State University\\
{\small \tt {\{adnan.qidwai@student,srija.mukhopadhyay@research\}.iiit.ac.in}; preranak@udel.edu;}\\
{\small \tt danroth@seas.upenn.edu; vgupt140@asu.edu}
}

\begin{document}
\maketitle
\begin{abstract}
Accurate and complete product descriptions are crucial for e-commerce, yet seller-provided information often falls short. Customer reviews offer valuable details but are laborious to sift through manually. We present \systemname: Product Review Attribute Insight Structuring Engine, a novel system that uses Large Language Models (LLMs) to automatically extract, compare, and structure insights from customer reviews and seller descriptions. \systemname provides users with an intuitive interface to identify missing, contradictory, or partially matching details between these two sources, presenting the discrepancies in a clear, structured format alongside supporting evidence from reviews. This allows sellers to easily enhance their product listings for clarity and persuasiveness, and buyers to better assess product reliability. Our demonstration showcases \systemname's workflow, its effectiveness in generating actionable structured insights from unstructured reviews, and its potential to significantly improve the quality and trustworthiness of e-commerce product catalogs.
\end{abstract}

\section{Introduction}
\label{sec:introduction}

In the rapidly expanding e-commerce landscape, platforms like Amazon heavily rely on detailed product descriptions to drive purchasing decisions \cite{vandic2018framework} and build customer trust \cite{reibstein2002attracts}. However, seller-provided descriptions frequently suffer from incompleteness or inaccuracies. While customer reviews often contain rich, factual information about product attributes and performance \cite{askalidis2016value}, manually extracting these details and reconciling them with seller descriptions is tedious and error-prone \cite{hu2004mining}. This gap highlights the need for automated tools to bridge information sources.

\begin{figure}[h!]
    \centering
    \includegraphics[width=0.85\linewidth]{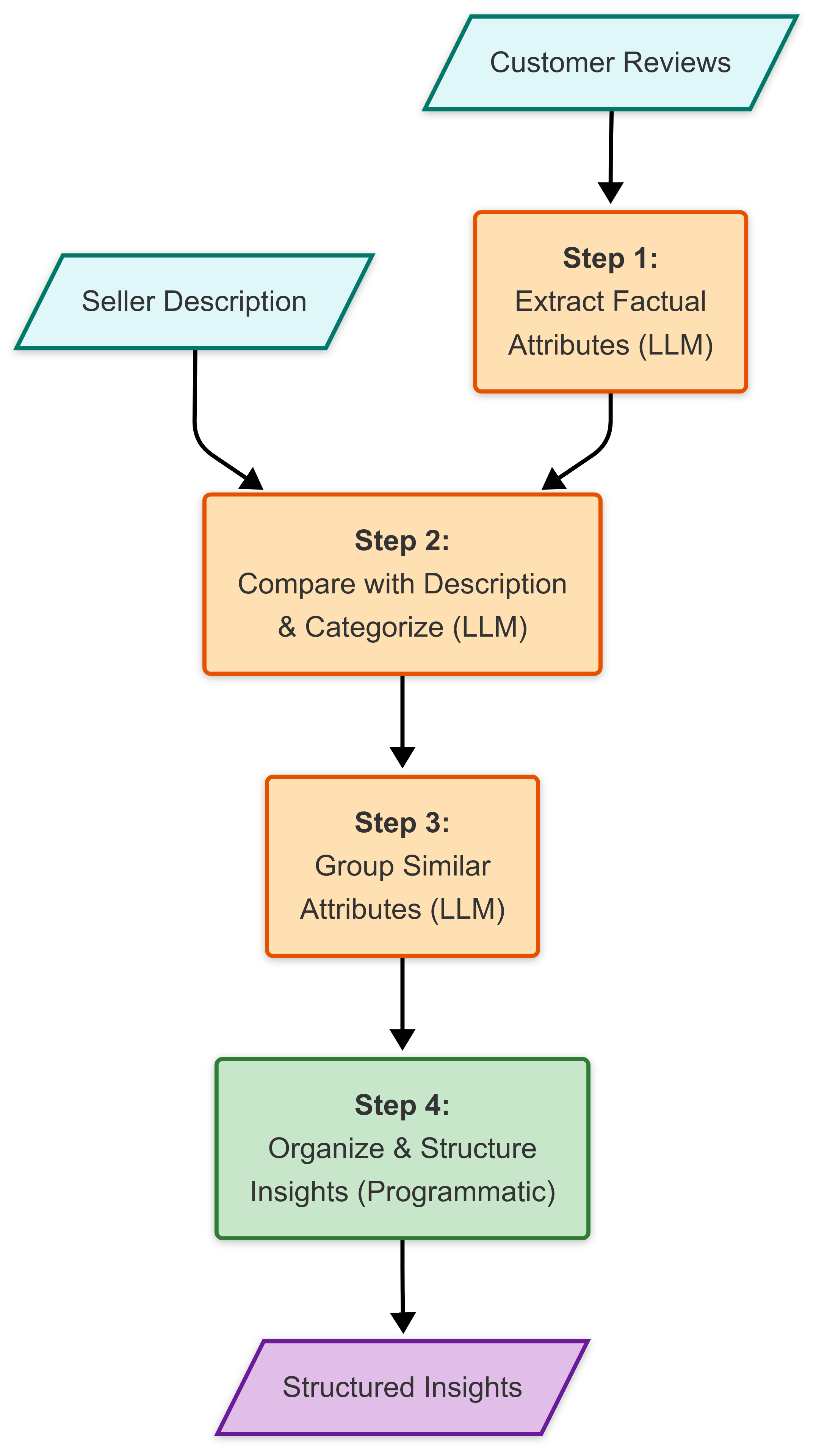}
    \vspace{-0.5em}
    \caption{End-to-End Pipeline of \systemname for Attribute Extraction and Structuring}
    \label{fig:architecture}
    \vspace{-1.75em}
\end{figure}

Recent advances in LLMs, with their proficiency in natural language understanding and generation \cite{roumeliotis2024llms, zhou2023leveraging, soni2023large}, offer a powerful means to address this challenge. Building on this potential, we developed \systemname, an interactive system designed to automatically enhance product descriptions using insights obtained from customer reviews.

\systemname's LLM-driven pipeline: (1) \textbf{Extracts} factual attributes from reviews (filtering opinions); (2) \textbf{Compares} attributes to seller descriptions; (3) \textbf{Categorizes} discrepancies (Missing, Contradictory, Partially-matching) with justifications; and (4) \textbf{Structures} findings for action. This allows users to quickly identify areas where product descriptions can be improved for accuracy and completeness. 
Our main contributions are:
\vspace{-0.3em}
\begin{enumerate}
    \item \textbf{The PRAISE System:} A novel, publicly accessible system demonstrating the use of LLMs for structured comparison of product descriptions and reviews.
Refer to Figure \ref{fig:architecture} for the complete system pipeline.     
    \vspace{-0.3em}
    \item \textbf{Evaluation and Insights:} An analysis of the system's performance, highlighting its strengths and current limitations in processing real-world e-commerce data.
\end{enumerate}
\vspace{-0.3em}


We invite readers to explore PRAISE’s capabilities through the following resources:
\begin{itemize}
\setlength{\itemsep}{0pt}
  \setlength{\parskip}{0pt}
  \item \textbf{Project Page:} \href{https://project-praise.github.io}{project-praise.github.io}
  \item \textbf{Demo Video:} \href{https://project-praise.github.io/demo/}{project-praise.github.io/demo/}
  \item \textbf{Try It Out:} \href{https://project-praise.github.io/tryout/}{project-praise.github.io/tryout/}
\end{itemize}
The system showcases a practical application of LLMs for tangible improvements in e-commerce information quality.

\looseness -1

\section{Related Work}
\label{sec:related_work}
Analyzing customer reviews for insights like sentiment and feature extraction has a rich history \citep{10.1145/1014052.1014073, popescu, mabrouk2021seopinion}. The advent of LLMs has significantly advanced capabilities in processing review data for tasks such as description generation, product categorization, and search refinement \citep{liu2023summary, roumeliotis2024llms, wang2024towards, choudhary2024interpretable}. Techniques like prompt engineering are vital for optimizing LLM outputs for specific tasks like information extraction and bias mitigation \citep{marvin2023prompt, 10.1055/a-2264-5631, wang}. While these works provide foundational techniques, PRAISE distinguishes itself by implementing a \textit{structured comparison} pipeline specifically designed to identify and categorize discrepancies (\textit{missing}, \textit{contradictory}, \textit{partial}) between review facts and seller descriptions, presenting them in an actionable format through an interactive system. Our focus is on demonstrating this end-to-end system for refining product catalog quality.

\section{The PRAISE System: Architecture and Workflow}
\label{sec:system}

PRAISE employs a multi-step pipeline, primarily leveraging LLMs, combined with programmatic orchestration to enrich product descriptions. The system processes customer reviews to extract pertinent descriptive information and systematically compares it against the seller-provided description. Our approach utilizes the advanced language understanding capabilities of LLMs for analysis, leveraging their ability to generate responses adhering to predefined structured formats (like JSON) where applicable, while integrating programmatic steps for structuring and organizing the results effectively.
The following steps detail this workflow:

\begin{figure}[h!]
    \centering
    \includegraphics[width=\linewidth]{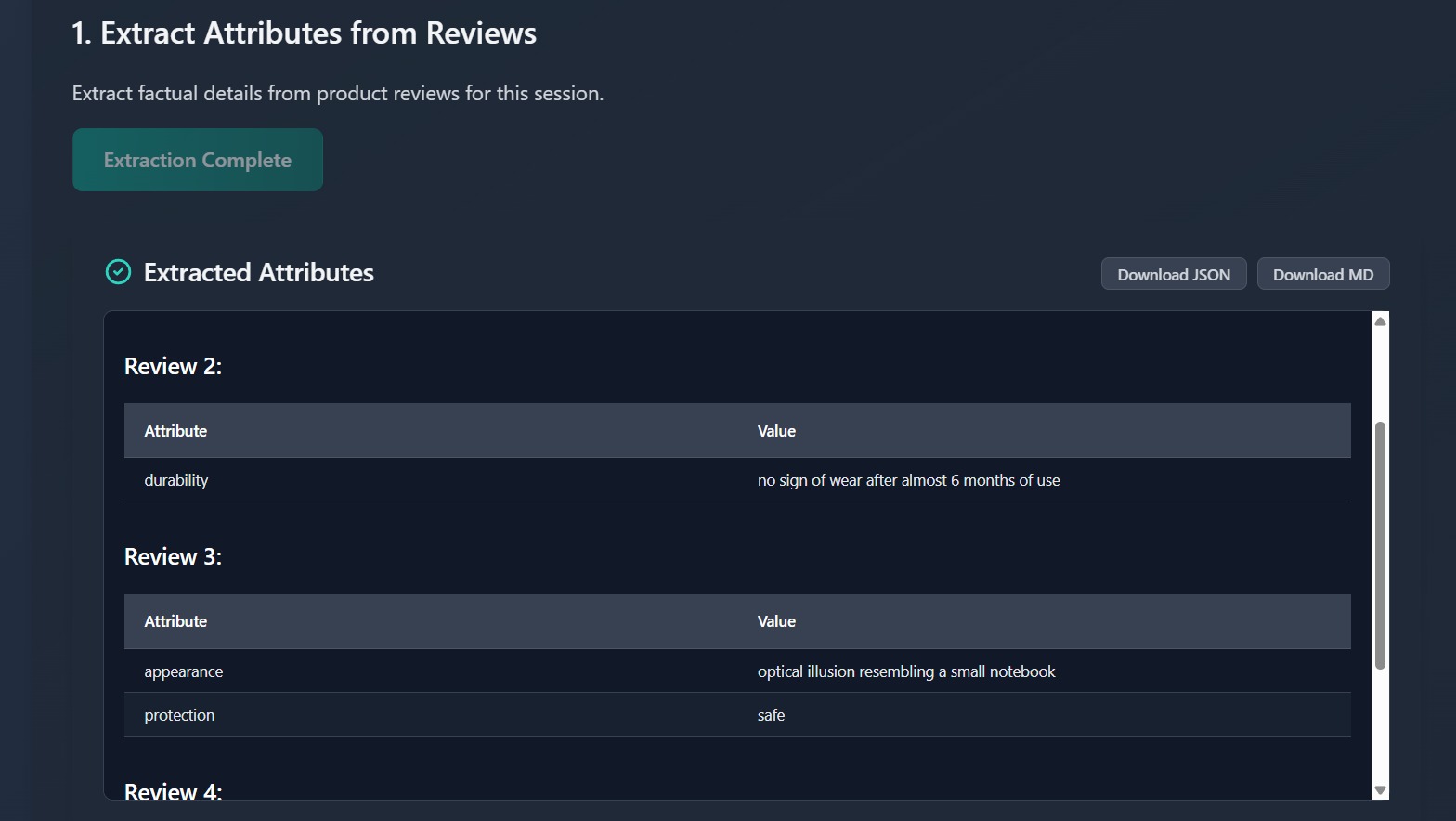}
    \vspace{-1em}
    \caption{Attribute extraction from product reviews (Step 1 of \systemname)}
    \label{fig:labelstep1}
    \vspace{-1em}
\end{figure}

\paragraph{Step 1: Extracting Descriptive Details from Reviews.}
The initial step focuses on analyzing each customer review individually to identify and isolate objective, factual information about the product (Figure \ref{fig:labelstep1}). An LLM is guided to distinguish these descriptive details (like materials, dimensions, or specific features) from subjective opinions, personal anecdotes, or irrelevant commentary. The core purpose is to filter out noise and retain only the verifiable, product-specific facts mentioned by reviewers. The output of this step is a collection of factual attribute-value pairs derived from each processed review.

\begin{figure}[h!]
    \centering
    \includegraphics[width=\linewidth]{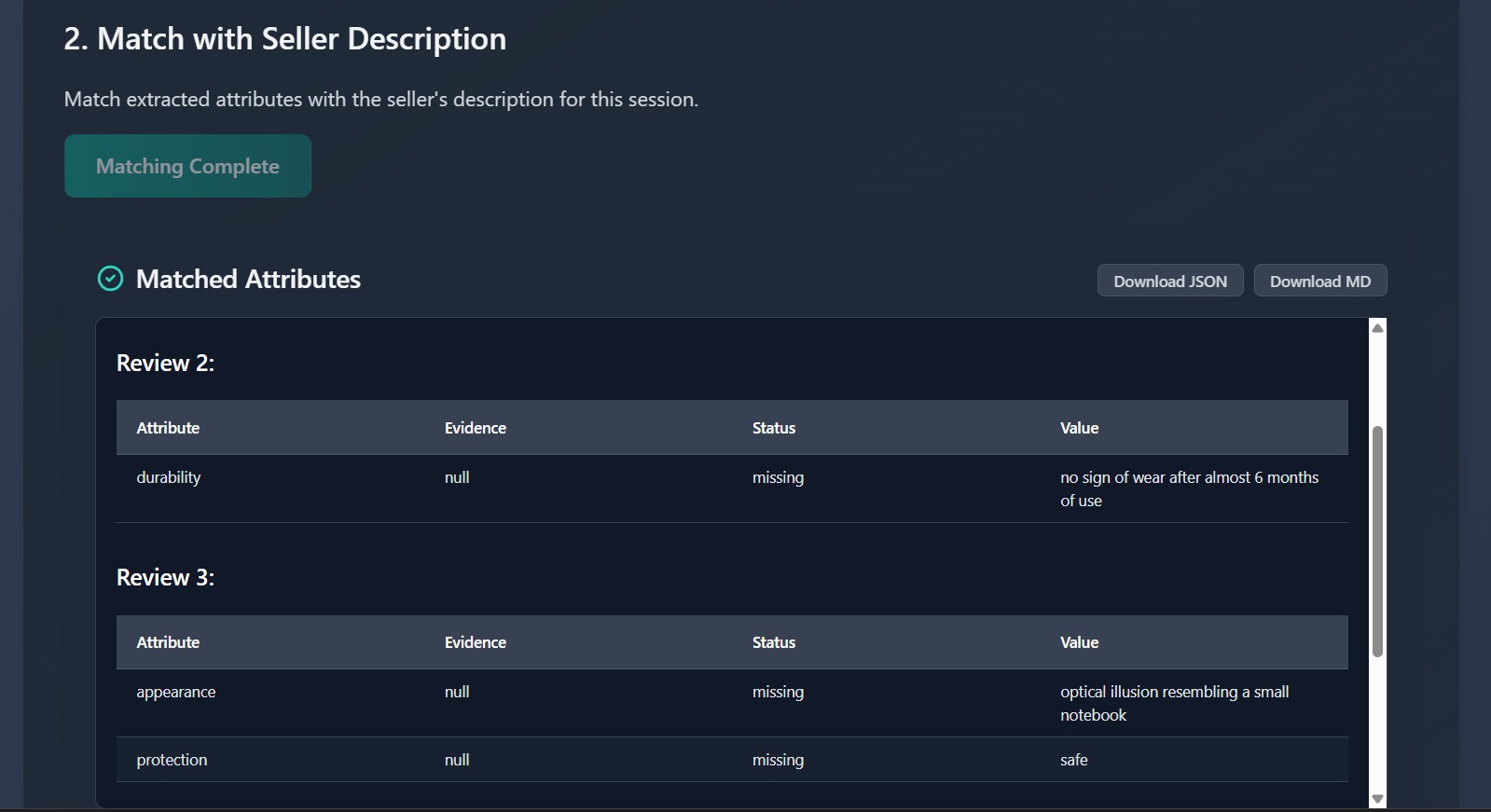}
    \vspace{-1em}
    \caption{Matching Extracted Attributes with Seller Descriptions (Step 2 of PRAISE)}
    \label{fig:labelstep2}
    \vspace{-1em}
\end{figure}

\paragraph{Step 2: Comparison with Seller Description and Categorization.}
Next, the system takes the factual attributes corresponding to each review extracted in the previous step and performs a comparative analysis against the official seller-provided product description (Figure \ref{fig:labelstep2}). An LLM examines each attribute derived from the reviews, determines its presence and consistency within the seller's text, and assigns a category based on the comparison. The key categories identify whether the information from the review is \textit{Missing} from the seller description, \textit{Contradictory} to it, \textit{Matching}, or only \textit{Partially-matching}. The process also includes providing reasoning or evidence from the seller's description where applicable. The outcome is a structured set of comparison results for each review, detailing how the factual points align or conflict with the seller's claims.

\begin{figure}[h!]
    \centering
    \includegraphics[width=\linewidth]{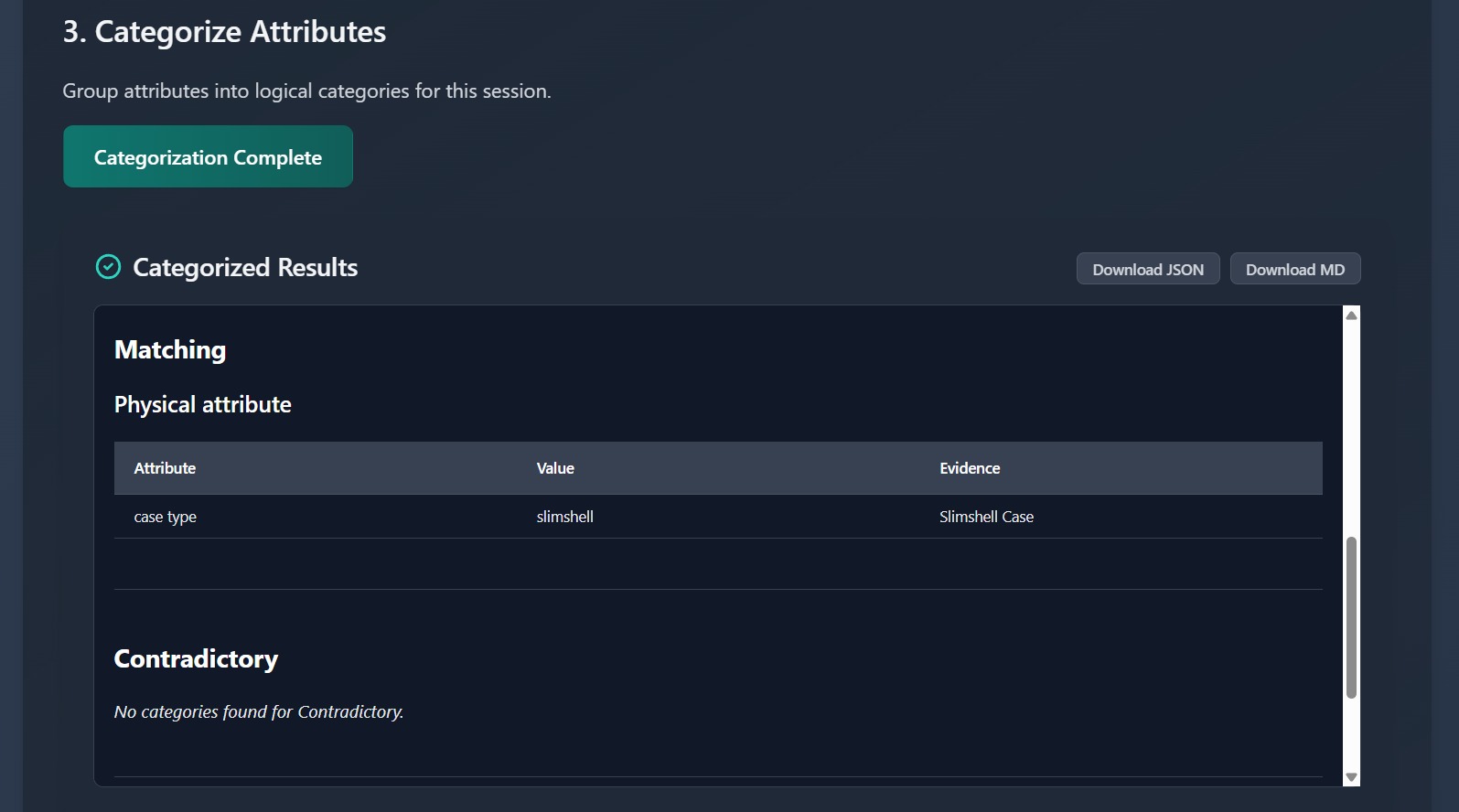}
    \vspace{-1em}
    \caption{Grouping and Structuring Attributes into Logical Categories (Steps 3 \& 4 of PRAISE)}
    \label{fig:groupingstep34}
    \vspace{-1em}
\end{figure}

\paragraph{Step 3: Grouping of Similar Attributes.}
To improve the organization of the findings, this step focuses on categorizing the diverse attributes identified across all reviews (Figure \ref{fig:groupingstep34}). Based on the attribute names (like `weight', `color', `battery life'), an LLM assigns each unique attribute to a broader, intuitive category (such as ``Physical Attributes'', ``Appearance'', ``Performance''). This grouping is based on the semantic similarity of the attribute concepts themselves, rather than their specific values, aiming for a generalized and user-friendly classification. The result of this step is a mapping that assigns a logical category to each type of attribute encountered.
\looseness -1

\paragraph{Step 4: Organizing and Presenting Structured Insights.}
In the final step, the system consolidates the comparison results from Step 2 and applies the category labels generated in Step 3. It programmatically structures this information through a \textit{rule-based} method, primarily highlighting the actionable insights – instances where review information was \textit{Missing}, \textit{Contradictory}, or \textit{Partially-matching} compared to the seller description. The findings are organized logically, grouped first by the comparison status and then by the attribute categories. This produces a final, structured output that presents the key discrepancies and alignments in an easy-to-navigate format, allowing users to quickly understand which aspects of the product description may require revision or verification based on customer feedback.

\paragraph{Pipeline Efficiency, Cost, and Optimization.}
The system's architecture is designed to balance performance with output quality. For a product with $R$ reviews, the pipeline makes approximately $2R+1$ LLM API calls, with the review-level extraction (Step 1) and comparison (Step 2) tasks being highly parallelizable. We used Python's \texttt{ThreadPoolExecutor} to execute these steps concurrently, significantly reducing wall-clock time. Robust error handling and retry logic are implemented for each LLM interaction to ensure resilience.
\looseness -1

This modular, multi-step design has direct implications for operational cost. While the linear scaling with the number of reviews is predictable, the cost can become a factor for products with extensive feedback. This design represents a deliberate trade-off: as demonstrated in our ablation analysis (Section \ref{sec:ablation_baseline}), aggregating steps into a single, cheaper API call (our end-to-end baseline) led to a significant degradation in quality, with higher rates of hallucination and incorrect categorization. The higher call volume of our modular pipeline is therefore a necessary investment to achieve reliable and actionable insights.

For production-scale deployment, several optimizations could further mitigate these costs without compromising quality:
\begin{itemize}
    \item \textbf{Model Tiering:} A tiered strategy could employ a powerful model (e.g., Gemini Pro) for the nuanced extraction and comparison tasks, while using a smaller, faster, and more cost-effective model (like the Gemini Flash model used in our experiments) for the simpler attribute grouping task (Step 3).
    \item \textbf{Caching:} Implementing a caching layer to store results for previously processed reviews and products would eliminate redundant API calls and computations entirely.
\end{itemize}
Thus, while our current implementation prioritizes demonstrative clarity and accuracy, a clear path exists for optimizing its cost-efficiency for large-scale use.
\paragraph{Implementation and Accessibility:}
Our public demonstration of \systemname is powered by the Gemini API. To ensure the system remains freely accessible for experimentation, users are required to provide their own Gemini API key, which is available at no cost and includes a generous free usage tier. To ensure user security, the provided key is handled exclusively on the server-side for the duration of the API calls and is never permanently stored or exposed to the client. This approach balances the practicalities of hosting a public LLM-powered demo with user accessibility and security.

\section{Methodology and Evaluation}
In this section, we describe the methodology behind the PRAISE system, detailing its multi-step pipeline for extracting, comparing, and structuring product insights using LLMs. 

\paragraph{Experimental Setup.}
To evaluate the performance of the PRAISE pipeline, we generated outputs using the implementation described in Section~\ref{sec:system}, primarily using the Gemini 2.0 Flash model. We selected Gemini 2.0 Flash for its strong balance of performance, inference speed, and cost-effectiveness, making it suitable for a scalable, interactive system. The evaluation dataset was derived from Amazon Reviews \citep{hou2024bridging}, encompassing 9 diverse product categories which included appliances, arts and crafts, beauty, books, CDs and vinyl records, cell phone accessories, clothing, shoes and jewelry, digital music and electronics. We selected around 10 products per category, with each product containing 7-9 reviews. We manually selected the reviews to make sure they had both opinions and descriptive details. This helped us accurately test how well our proposed system works.

\paragraph{Evaluation Protocol.}
A panel of three research team members manually verified the outputs of the pipeline’s core LLM-driven stages: Step 1 (Attribute Extraction), Step 2 (Review-Seller Comparison), and Step 3 (Attribute Grouping). Evaluators used the original reviews and seller descriptions as ground truth and followed detailed, pre-defined rubrics to ensure consistent assessment.


Each identified error was counted as one point, enabling a quantitative analysis of error frequencies and types. Annotators followed a detailed evaluation rubric to identify specific error categories across all stages of the pipeline. In Step 1 (Extraction), common issues included incorrect attribute–value extraction, failure to filter out opinions, inclusion of irrelevant (non-product) information, and omission of valid attributes. In Step 2 (Comparison), evaluation focused on the correctness of the assigned status—Missing, Matching, Contradictory, or Partially Matching—as well as the validity of the accompanying justifications, with particular attention to misclassifications between these categories. In Step 3 (Grouping), errors involved inappropriate category naming, incorrect assignment of attributes to categories, and issues with grouping granularity, including both over-splitting and under-splitting of semantically related attributes.


\section{Results and Analysis}

\subsection{Key Observations from Error Analysis}
The quantitative evaluation provided specific insights into the performance and challenges of each pipeline stage using the Gemini 2.0 Flash model.

\begin{figure}[h!]
    \centering
    \includegraphics[width=0.5\textwidth]{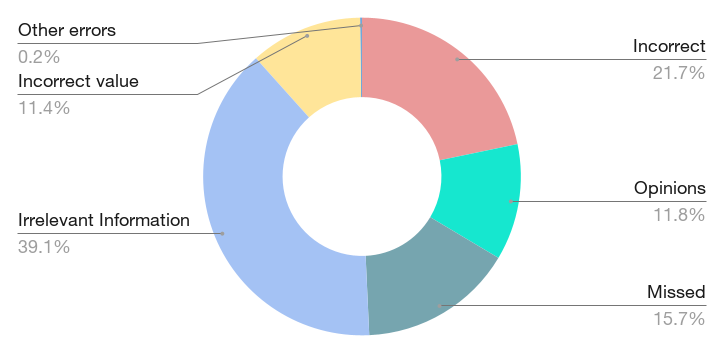}
    \caption{Distribution of Error Types in Step 1}
    \label{fig:errordist1}
\end{figure}

\paragraph{Step 1 (Extraction):} This initial step, focused on extracting factual attributes from raw reviews, exhibited the highest error frequency.
\begin{itemize}
    \item \textit{Strengths:} Despite these challenges, the system successfully extracted many basic factual attribute-value pairs, forming the necessary input for subsequent steps. The low count for `Other errors' (2) suggests the defined rubric categories comprehensively captured the types of issues encountered.
    
    \item \textit{Weaknesses:} The most prominent issue was the inclusion of \texttt{Irrelevant Information} (403 instances), where the model struggled to differentiate product-specific details from user context or opinions disguised as facts. Assigning accurate attribute names also proved difficult, leading to numerous \texttt{Incorrect Normalization} errors (224). 
\end{itemize}
Figure \ref{fig:errordist1} demonstrates the major categories showcasing the errors made by the system, as found out by our robust evaluation methodology.

\begin{figure}[h]
    \centering
    \includegraphics[width=0.5\textwidth]{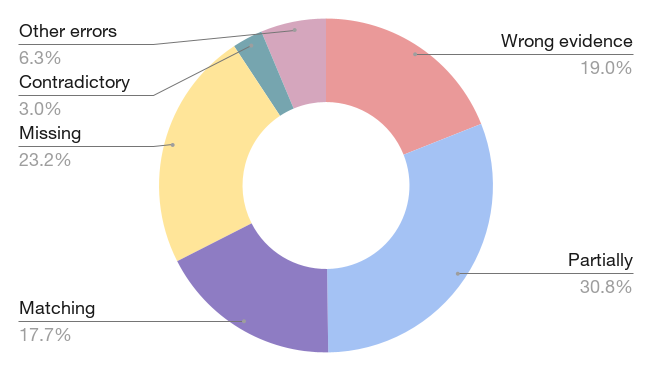}
    \vspace{-0.5em}
    \caption{Error Distribution in Step 2—Matching Extracted Attributes with Seller Descriptions}
    \vspace{-0.5em}
    \label{fig:errordist2}
\end{figure}

\textbf{Step 2 (Comparison):} Figure \ref{fig:errordist2} shows the errors encountered during Step 2. This step compared extracted attributes to the seller description, showing a pretty low number of errors (approx. 237 total) compared to extraction. This also shows the ability of the system to stop errors from cascading onto further steps, enabled by our robust prompt engineering. 
\begin{itemize}
    \item \textit{Strengths:} The system performed considerably better here than in Step 1, successfully categorizing many attributes. Notably, the misclassification rate for \texttt{Contradictory} status was very low (7 instances), suggesting the model is relatively conservative or accurate when identifying direct conflicts, which is valuable for highlighting critical discrepancies.
    \looseness -1
    \item \textit{Weaknesses:} The primary difficulty lay in accurately classifying the relationship between review and description attributes. Misclassifications where the model incorrectly identified attributes as \texttt{Partially Matching} (73 instances) or \texttt{Missing} (55 instances) were most common, indicating struggles with nuanced semantic differences versus true absences.
\end{itemize}

\begin{figure}[h]
    \centering
    \includegraphics[width=0.5\textwidth]{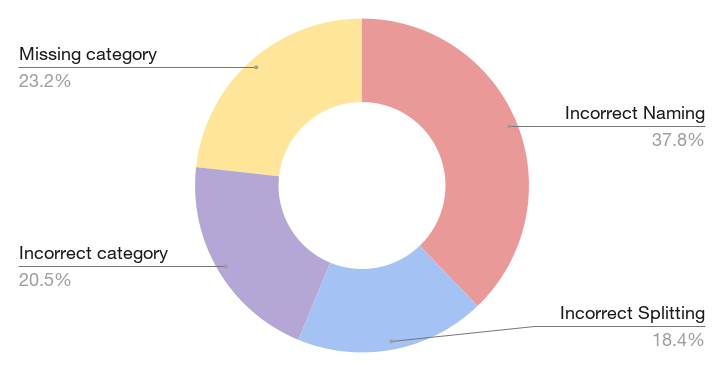}
    \vspace{-0.5em}
    \caption{Breakdown of Error Types in Step 3--Attribute Grouping and Categorization}
    \label{fig:errordist3}
    \vspace{-0.5em}
\end{figure}

\textbf{Step 3 (Grouping):} Tasked with grouping related attributes based on their keys, this step was the most robust, exhibiting the fewest errors (approx. 187 total). This further shows how our system produces helpful end results. 
\begin{itemize}
    \item \textit{Strengths:} The relatively low overall error count suggests that grouping based primarily on attribute keys is an effective strategy for this LLM. It successfully organized the majority of attributes into reasonable clusters. The low count for 'Other errors' (2) again implies the rubric covered the main issues.

    \item \textit{Weaknesses:} The most frequent error was \texttt{Incorrect Category Naming} (70 instances), where the LLM generated labels that were not optimally descriptive or semantically appropriate for the grouped attributes. Issues with grouping granularity were also present, including \texttt{Missing Category} (under-splitting, 43 instances) where distinct concepts were wrongly merged, and \texttt{Incorrect Splitting} (over-splitting, 34 instances) where attributes were unnecessarily separated.
\end{itemize}

We show a more in-depth analysis of errors encountered during Step 3 in Figure \ref{fig:errordist3}. 

\subsection{Product-Wise Error Analysis}
The evaluation revealed significant variations in pipeline performance across different product categories, as measured by precision, recall, and F1 score. This indicates that the typical language used the reviews of products heavily influence the system's ability to accurately identify relevant attributes. Table \ref{tab:prec_recall} presents these detailed performance metrics for each category.

\begin{table}[ht]
\centering
\resizebox{\columnwidth}{!}{%
\begin{tabular}{l|c|c|c}
\hline
\textbf{Category} & \textbf{Precision} & \textbf{Recall} & \textbf{F1 Score} \\ \hline
Appliances & 0.41 & 0.84 & 0.55 \\
Arts and Crafts & 0.72 & 0.96 & 0.82 \\
Beauty & 0.6 & 0.99 & 0.75 \\
Books & 0.23 & 0.79 & 0.36 \\
CD Vinyl & 0.52 & 0.6 & 0.56 \\
Cell Phone and Accessories & 0.56 & 0.82 & 0.66 \\
Clothes, Shoes, Jewelery & 0.45 & 0.79 & 0.57 \\
Digital music & 0.46 & 0.71 & 0.56 \\
Electronics & 0.47 & 0.86 & 0.61 \\ \hline
\end{tabular}%
}
\vspace{-0.5em}
\caption{Performance Metrics by Product Category for Attribute Selection Using PRAISE}
\label{tab:prec_recall}
\vspace{-1.0em}
\end{table}

\paragraph{High-Performance Categories:}
Models performed the best in \textit{Arts and Crafts} with the highest F1 score (0.82), driven by excellent recall (0.96) and good precision (0.72). This suggests the system is highly effective at identifying the vast majority of true attributes for this category, while also maintaining a relatively low rate of incorrectly selecting non-attributes. 

A similar performance was noticed for \textit{Beauty} products with an F1 score of 0.75. Notably, this category achieved near-perfect recall (0.99), indicating the system rarely misses a relevant beauty attribute. However, its precision (0.60) is moderate, suggesting that while comprehensive, the system may also select a fair number of terms that are not true attributes, possibly due to the highly descriptive and often subjective language in beauty reviews (e.g., "silky," "glow," "subtle scent").

\paragraph{Challenging Categories:}
This group, consisting \textit{Clothes, Shoes, and Jewelry} (F1 0.57), \textit{CD \& Vinyl} (F1 0.56), \textit{Digital Music} (F1 0.56), \textit{Appliances} (F1 0.55), and most notably \textit{Books} (F1 0.36), caused significant hurdles in attribute selection, primarily due to low precision. 

For categories like \textit{Appliances} (P 0.41, R 0.84) and \textit{Clothes, Shoes, and Jewelry} (P 0.45, R 0.79), good recall was undermined by very low precision where the system identified most true attributes but also incorrectly selected many non-attribute terms due to complex technical/usage details (Appliances) or subjective and overly descriptive language. 

The music categories (\textit{Digital Music}: P 0.46, R 0.71; \textit{CD \& Vinyl}: P 0.52, R 0.60) showed modest overall scores, struggling with precise extraction despite standardized metadata, likely because reviews often prioritize opinions over discrete factual features. \textit{Books} was the most problematic, with acceptable recall (0.79) but exceptionally low precision (0.23). This starkly indicates that the high volume of subjective commentary, thematic discussions, and fewer distinct factual attributes in book reviews makes it exceedingly difficult for the system to distinguish true attributes from textual noise, leading to a very high rate of false positives.

\paragraph{Overall Implications for Attribute Selection:}
The category-specific analysis reveals distinct performance profiles. While some categories (\textit{Arts and Crafts, Beauty}) achieve high recall, effectively identifying most true attributes, precision is a primary challenge across many others. This is particularly true for categories characterized by complex technical language (e.g., \textit{Appliances, Electronics}), high subjectivity (e.g., \textit{Beauty, Books}), or a lot of descriptive text (e.g., \textit{Clothes}). The starkly low precision for \textit{Books}, despite its standardized metadata, illustrates how a high volume of subjective or descriptive text can severely affect accurate attribute selection. Future efforts must prioritize enhancing the system's discrimination between true attributes and textual noise, particularly for these low-precision categories.

\section{Baseline and Ablation Analysis}
\label{sec:ablation_baseline}

\begin{figure}[h!]
    \centering 
    \includegraphics[width=\linewidth]{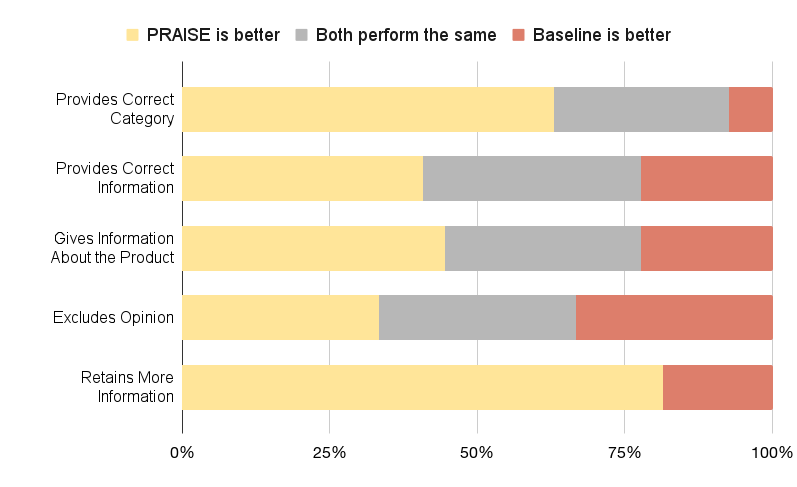} 
    \vspace{-1.5em}
    \caption{Comparison of the full PRAISE pipeline against the single-prompt Baseline.}
    \label{fig:ab0}
    \vspace{-1.0em}
\end{figure}

\begin{figure}[h!]
    \centering 
    \includegraphics[width=\linewidth]{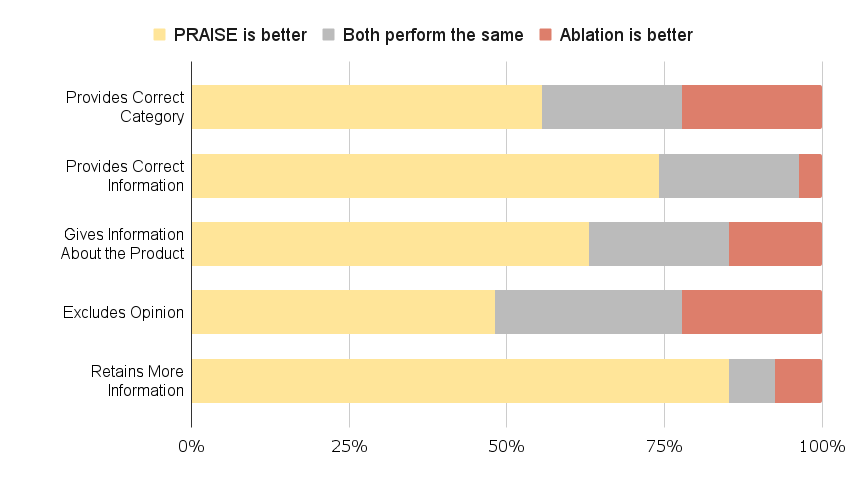} 
    \vspace{-1.75em} 
    \caption{Comparison of the full PRAISE pipeline against the Ablated System (Step 1 + Direct Prompt).}
    \label{fig:ab1}
    \vspace{-1.0em}
\end{figure}

To evaluate the contribution of our multi-step pipeline design, we conducted comparative analyses against two simpler approaches. We assessed performance based on several criteria reflecting the quality of the final structured output: the ability to retain important product details, exclude subjective opinions, maintain focus on product-specific information, and accurately categorize information (e.g., as missing or contradictory). Performance differences are illustrated by comparing counts across these criteria, as shown in Figure \ref{fig:ab0} and Figure \ref{fig:ab1}.

\paragraph{Baseline Comparison: End-to-End Prompting.}
We compared PRAISE against a \textit{Baseline} system. This baseline used a single, direct prompt instructing the LLM to perform the entire task end-to-end – taking raw reviews and seller description as input and generating the final structured output format without the intermediate processing steps defined in our pipeline. This comparison establishes a benchmark against a less structured, single-shot approach.

As shown in Figure \ref{fig:ab0}, the full PRAISE pipeline consistently outperformed the baseline across nearly all evaluated metrics, particularly in areas like correct categorization and opinion exclusion. This demonstrates the significant value added by the structured intermediate steps in maintaining accuracy and information fidelity compared to direct end-to-end generation.

\paragraph{Ablated System Comparison: Isolating Later Stages.}
Second, we evaluated an \textit{Ablated System}. This approach performed only Step 1 of our pipeline (structured attribute extraction) and then used a direct prompt to generate the final categorized output directly from these extracted attributes, bypassing the explicit comparison, grouping, and organization stages (Steps 2-4). This comparison isolates the contribution of these later structured processing steps, given the initial structured extraction.
\looseness -1

\vspace{-1.75em}

Figure \ref{fig:ab1} clearly indicates that the full PRAISE pipeline significantly outperformed this ablated system across all evaluation criteria. While structured extraction (Step 1) provides a better starting point than raw text (as suggested by the Baseline comparison), the results here highlight that the subsequent dedicated steps for comparison (Step 2), grouping (Step 3), and organization (Step 4) are crucial for achieving the highest level of accuracy and generating the most reliable structured insights. Together, these comparisons validate the effectiveness of our complete multi-step pipeline design.

\section{Model Effectiveness Comparison}
\paragraph{Multi-Step vs Automatic Prompt.}
To further validate our method, we conducted supplementary analyses for establishing its credibility and demonstrating its advantages over alternative approaches. We first tested a single, all-in-one prompt to complete the entire task. This helped us show that our multi-step approach is more accurate, consistent, and easier to understand. We then bench marked against automatic prompting. We used the intent-based prompt calibration technique outlined in \citep{levi2024intentbased} to automatically calibrate prompts for the task of catalog expansion. This comparison highlights how our well-designed prompts and step-by-step approach lead to better performance.
\looseness -1

When attempting to execute the entire process in a single step, we observed a significant increase in hallucination from the model. This suggests that the model struggled to differentiate between the various stages of the task and consequently lost crucial information. Similarly, the automatic prompt generation approach yielded suboptimal results, likely due to the inherent complexity of the task and the lack of a well-defined evaluation metric for this specific application.

\paragraph{System License Details.} LLMs used in this study are licensed by their creators, while our platform uses the Apache 2.0 License. 
This license permits any use, distribution, modification, and commercial use of the software, including sublicensing and adding warranties.

\section{Conclusion}
Our work shows that LLMs can improve e-commerce product listings by integrating insights from customer reviews. We used a structured approach to filter out opinions and keep factual information, allowing us to accurately compare reviews with seller descriptions. This helped identify matches, gaps, and contradictions, and organize information into clear tables for easier analysis. Our results indicate that while LLMs are effective at summarizing and correcting spellings, they can struggle with filtering subjective opinions. Overall, LLMs are useful for improving product descriptions but have varying performance depending on the task, the model and the quality of reviews and descriptions provided.

Future work will focus on improving attribute relevance filtering and refining key-value alignment to enhance precision. Expanding attribute categorization and incorporating Q\&A data will boost insight quality and coverage. Finally, we plan to test the system across more product categories and extend support to multilingual inputs.

\section*{Limitations}
While PRAISE demonstrates strong performance in structuring review-based insights, it is limited by its exclusive reliance on customer reviews, which may vary in quality and clarity, leading to inconsistent extractions. The current system does not incorporate other user-generated content such as Q\&A pairs, which could enhance coverage. It is restricted to English-language inputs, reducing its applicability in multilingual settings. Evaluation depends on human judgment, which introduces subjectivity and limits scalability. The model may struggle with nuanced semantic distinctions, especially in subjective or mixed-content reviews, and can extract irrelevant or noisy information. Moreover, the use of a fixed prompt-based pipeline with a single LLM may constrain adaptability across diverse product categories. Finally, the absence of end-user feedback mechanisms limits our ability to assess real-world utility and usability.

\section*{Ethics Statement}
PRAISE employs large language models (LLMs) to extract factual information from customer reviews, introducing several ethical considerations. While the system incorporates step-wise prompting and structured evaluation to mitigate common failure modes, LLMs remain susceptible to producing biased or inaccurate outputs due to limitations in their training data. The reviews processed are publicly available and free of personally identifiable information; however, any future extensions involving private or sensitive data must ensure robust privacy protections. The reliance on proprietary models constrains transparency and interpretability, which we partially address through systematic documentation and error analysis. The system may also be vulnerable to misuse, such as selectively emphasizing favorable attributes or suppressing critical ones. To reduce this risk, PRAISE is explicitly designed to extract verifiable content and highlight missing or contradictory details. The system is released under the Apache 2.0 license, and users must supply their own Gemini API keys, which are not stored or logged. Responsible deployment of PRAISE requires human oversight to safeguard fairness, ensure accountability, and prevent potential misuse.

\section*{Acknowledgement}
This work was primary sponsored by the Army Research Office under Grant Number W911NF-20-1-0080. The views and conclusions expressed herein are those of the authors and do not necessarily reflect the official policies, either expressed or implied, of the Army Research Office or the U.S. Government. The U.S. Government is authorized to reproduce and distribute reprints for governmental purposes notwithstanding any copyright notice herein. This research was also partially supported by ONR Contract N00014-19-1-2620.

We thank the CogComp group (UPenn) for their insights; Mahika Vajpeyi (UPenn) for initial discussions; and Preethi Kallimuddanahalli Suresh (ASU) for help with dataset wrangling. We are also grateful to our anonymous reviewers for their thoughtful feedback and to the CoRAL Lab (ASU) for computational support. Special thanks to our lab cat, Coco, for keeping Prof. V.G. creatively unhinged during deadlines.

\bibliography{main}
\bibliographystyle{aclnatbib}



\end{document}